\documentclass[10pt,twocolumn,letterpaper]{article}

\usepackage{cvpr}              % To produce the CAMERA-READY version
\usepackage{array}
\newcolumntype{C}[1]{>{\centering\arraybackslash}p{#1}}

\usepackage[dvipsnames]{xcolor}

\definecolor{cvprblue}{rgb}{0.21,0.49,0.74}
\usepackage[pagebackref,breaklinks,colorlinks,citecolor=cvprblue]{hyperref}

\def\paperID{8} % *** Enter the Paper ID here
\def\confName{CVPR}
\def\confYear{2024}

\title{Multi-Frame, Lightweight \& Efficient Vision-Language Models for Question Answering in Autonomous Driving}

\author{Akshay Gopalkrishnan\\
{\tt\small agopalkr@ucsd.edu}
\and
Ross Greer\\
{\tt\small regreer@ucsd.edu}
\and
Mohan Trivedi\\
{\tt\small mtrivedi@eng.ucsd.edu}
}

\begin{document}
\maketitle
\begin{abstract}
Vision-Language Models (VLMs) and Multi-Modal Language Models (MMLMs) have become prominent in autonomous driving research, as these models can provide interpretable textual reasoning and responses for end-to-end autonomous driving safety tasks using traffic scene images and other data modalities. However, current approaches to these systems use expensive large language model (LLM) backbones and image encoders, making such systems unsuitable for real-time autonomous driving systems where tight memory constraints exist and fast inference time is necessary. To address these previous issues, we develop EM-VLM4AD, an efficient, lightweight, multi-frame vision language model which performs Visual Question Answering for autonomous driving. In comparison to previous approaches, EM-VLM4AD requires at least 10 times less memory and floating point operations, while also achieving higher CIDEr and ROUGE scores than the existing baseline on the DriveLM dataset. EM-VLM4AD also exhibits the ability to extract relevant information from traffic views related to prompts and can answer questions for various autonomous driving subtasks. We release our code to train and evaluate our model \href{https://github.com/akshaygopalkr/EM-VLM4AD/tree/main}{here}.
\end{abstract}    
\section{Introduction}
\label{sec:intro}

\indent Vision-Language Models (VLMs) have emerged as powerful tools that possess holistic knowledge to solve tasks at the intersection of vision and language. This makes them a promising asset in autonomous driving (AD), allowing for a driver to interact with the VLM which can provide interpretable language representations of various driving safety tasks. Furthermore, VLMs can serve as end-to-end autonomous driving systems, eliminating integration and propagating errors between separate models specializing in specific sub-tasks of autonomous driving such as perception \cite{greer2023robust, greer2024patterns, greer2024and} and trajectory planning \cite{messaoud2021trajectory}. These potential benefits have propelled the development of many vision-language models and multimodal language models tailored for autonomous driving applications \cite{chen2023driving, touvron2023llama, xu2023drivegpt4, mao2023gpt, sima2023drivelm}. These models cover various aspects of autonomous driving including closed-loop control, perception tasks, and traffic agent behavior analysis. 

Typically, the process in a VLM is the following: vision and text features are encoded separately, then fused together through a concatenation or projection layer, and then finally fed into an LLM to output some probability distribution over the vocabulary \cite{wu2024mivc}. While generating text embeddings is relatively low-cost, the LM and image embeddings can often entail high computational costs. In real-time systems such as autonomous driving, prioritizing the development of VLMs with efficient inference times is crucial for practical deployment in vehicles. However, current research in applying multimodal language models to autonomous driving predominantly use large models such as BLIP-2 \cite{li2023blip}, GPT 3.5 \cite{mao2023gpt}, and LLaMA-7b \cite{touvron2023llama}, all of which contain over one billion parameters. Consequently, these models require expensive hardware and longer inference times, limiting their potential to be applied in current vehicles and accessibility for researchers with limited computational resources. 

This paper focuses on the development of lightweight vision-language models with less than one billion parameters than can accurately and efficiently answer questions related to autonomous driving safety tasks. We develop the model EM-VLM4AD: Efficient, Multi-Frame Vision-Language Model for Autonomous Driving. We use the DriveLM dataset \cite{sima2023drivelm}, which offers real, multi-view traffic scene images paired with question/answer pairs to train this model. Our contributions are as follows:
\begin{itemize}
    \item We develop an efficient, smaller vision-language model EM-VLM4AD that consumes at least \textbf{10x} less memory and floating point operations (FLOPs) than current AD-VLMs, and can also respond to questions conditioned on multiple frames.
    \item We explore two different lightweight LM backbones for EM-VLM4AD: a finetuned Text-to-Text Transfer Transformer (T5) Base LM and an 8-bit quantized T5-Large LM finetuned using low-rank adaptation (LoRA) \cite{hu2021lora}.
    \item We compare our model efficiency and performance on BLEU-4 (Bilingual Evaluation Understudy), CIDEr (Consensus-based Image Description Evaluation), ROUGE-L (Recall-Oriented Understudy for Gisting Evaluation), and METEOR (Metric for Evaluation of Translation with Explicit Ordering) to the baseline for the DriveLM dataset \cite{sima2023drivelm}, demonstrating stronger performance in ROUGE-L and CIDEr even with superior efficiency using a much smaller model. 
\end{itemize}

\section{Related Research}
\label{sec:related research}

\subsection{Vision-Language Models}
Initially designed to operate on sequence data, Transformers \cite{vaswani2017attention} achieved state-of-the-art performance for natural language processing tasks. This propelled the development of Large Language Models, which learn general statistical properties of language through pretraining Encoder \cite{devlin2018bert}, Encoder-Decoder \cite{raffel2023exploring}, and Decoder \cite{radford2019language, touvron2023llama, almazrouei2023falcon} Transformer architectures on a large corpus of tokens. These pre-trained models can then be finetuned for downstream, more specialized language tasks. Dosovitskiy et al. \cite{dosovitskiy2020image} introduced the application of Transformers to image tasks with the Vision Transformer (ViT), which converts images into a sequence representation of image patches that can be processed by Transformers. Vision-Language Models bridge the gap between LLMs and Vision Transformers, encoding images and text into a combined latent representation and then utilizing cross-modal pre-training tasks to learn text and image correlations. This general approach to multimodal learning has sparked a variety of vision-language models. Radford et al. \cite{radford2021learning} devise a pre-training task of matching text captions with images to develop CLIP, which learns state-of-the-art image representations and exhibits strong zero-shot transfer capabilities for many image classification tasks. BLIP-2 \cite{li2023blip} introduces a two stage pretraining process to train a Querying Transformer ``QFormer" that serves as a intermediary between a frozen image encoder and language model. This approach outperforms much larger vision-language models such as Flamingo \cite{alayrac2022flamingo} and is capable of zero-shot image-to-text generation. InstructBLIP \cite{dai2023instructblip} builds off BLIP-2 and is a general-purpose VLM that aggregates public vision-language datasets and transforms them into an instruction tuning format. The VLM most similar to the model introduced in this paper is VL-T5 \cite{cho2021unifying}, which extends a pre-trained T5 to learn to generate text labels conditioned on a combination of a text and image embedding. Using a pre-trained LLM as a framework for multi-modal tasks harnesses the text generation ability of these models, critical for the question-answering task in our research. Despite their strong performance across many tasks, deploying these large models, which often exceed one billion parameters, is difficult for real-time applications \cite{du2022survey}. Consequently, researching compression techniques like distillation \cite{li2023distilling, fang2021compressing}, quantization, and pruning is imperative to reduce VLM latency and computational costs. 

\subsection{Multimodal LLMs for Autonomous Driving}
While autonomous driving systems mainly use visual features, introducing linguistic features can enhance the interpretability of these systems and even help identify novel traffic situations \cite{greer2024towards}. This benefit has sparked research interest in integrating multimodal data to train language models to become autonomous driving agents. Chen et al. \cite{chen2023driving} design an architecture that fuses vectorized numeric modalities with a pretrained LLaMA-7b \cite{touvron2023llama} to solve Driving Question Answering tasks. Using a two-step training approach, they initially ground the vector representations into interpretable embeddings for the frozen LLaMA model, followed by finetuning the LLM with LoRA \cite{hu2021lora}. DriveGPT4 \cite{xu2023drivegpt4} also adopts LLaMA as a backbone LLM and CLIP as a visual encoder, using a traffic scene video and prompt text as input to generate answers and low-level vehicle control signals. To expand off the fixed and rigid QA labels from the BDD-X dataset \cite{kim2018textual}, DriveGPT4 is trained on instruction tuning data generated by ChatGPT/GPT4. DriveGPT4 only uses a single-view camera, which restricts it to questions involving a single view. Wang et al. \cite{wang2023drivemlm} introduce DriveMLM, which uses multi-view images, LiDAR Point Clouds, traffic rules, and user commands from a realistic simulator to perform closed-loop driving. This multimodal model is built from LLaMA-7B and ViT-g/14 as the image processor. Sha et al. \cite{sha2023languagempc} devise a chain-of-thought \cite{wei2023chainofthought} framework for driving scenarios using ChatGPT 3.5 to provide interpretable, logical reasoning for autonomous driving systems. Mao et al. \cite{mao2023gpt} also leverage the GPT-3.5 model to create a motion planner for autonomous vehicles. Their model, GPT-Driver, reformulates motion planning as a language modeling problem by representing planner inputs and outputs as language tokens. Recently, Sima et al. \cite{sima2023drivelm} released the DriveLM dataset, a Graph Visual Question Answering dataset that provides question-answer pairs related to perception, behavior, and ego-vehicle planning based off multi-view image data from the NuScenes dataset \cite{caesar2020nuscenes}. To introduce a baseline, Sima et al. finetune BLIP-2 \cite{li2023blip} for this novel dataset.

While these approaches provide valuable explainability for AD systems and exhibit strong performance for end-to-end tasks, all these models use LLMs with over one billion parameters (GPT 3.5, LLaMA, etc.) and expensive image encoders like CLIP and ViT-g/14. This makes them primarily suitable for offline scenarios where latency is not a priority, but not for online situations where real-time inference is paramount. 

\subsection{Multi-Image Vision-Language Models}
In the realm of autonomous driving, modalities beyond text and image such as LiDAR, radar, or video offer important features for many downstream tasks. However, most vision-language models are pre-trained for single-image single-text problems, making it unfeasible to directly input multiple images or modalities with a piece of text \cite{wu2024mivc}. Consequently, it is necessary to consolidate multiple modalities and text into a single embedding that can be used by a VLM. DriveGPT4 \cite{xu2023drivegpt4} encodes video input by pooling CLIP visual encodings of each video frame. DriveMLM's \cite{sima2023drivelm} multimodal tokenizer uses QFormer to embed video and LiDAR data, and then concatenates these embeddings with text and system message embeddings. Wu et al. \cite{wu2024mivc} find that using gated attention pooling across each individual image embedding helps introduce more non-linearity and extracts visual information across multiple images. Importantly, this gated attention method introduces a negligible amount of computational overhead, rendering it an ideal choice for our model to aggregate multi-view traffic scene images into a unified embedding. 
\begin{figure*}
    \centering
    \includegraphics[scale=0.8]{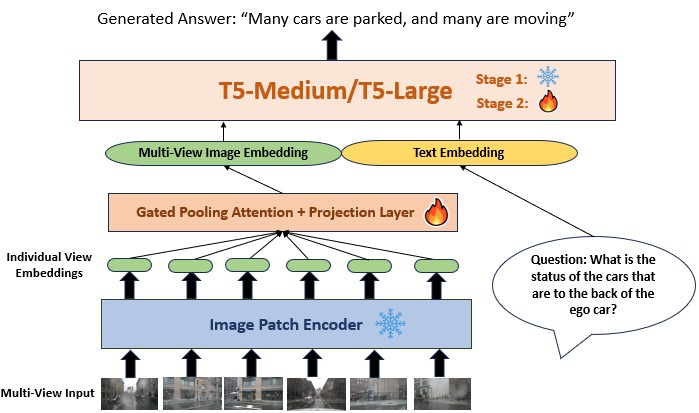}
    \caption{The diagram our model uses to respond to multi-view image input and question prompts. The T5 LM is frozen during Stage 1 of training so the image embedding network learns to align with the T5 embeddings. The image patch encoder is frozen throughout all stages of training, and the Gated Pooling Attention and Projection Layer is trained in both stages.}
    \label{fig:DriveLM-examples}
\end{figure*}

\begin{figure*}[hbt!]
\centering
    \includegraphics[width=0.8\textwidth]{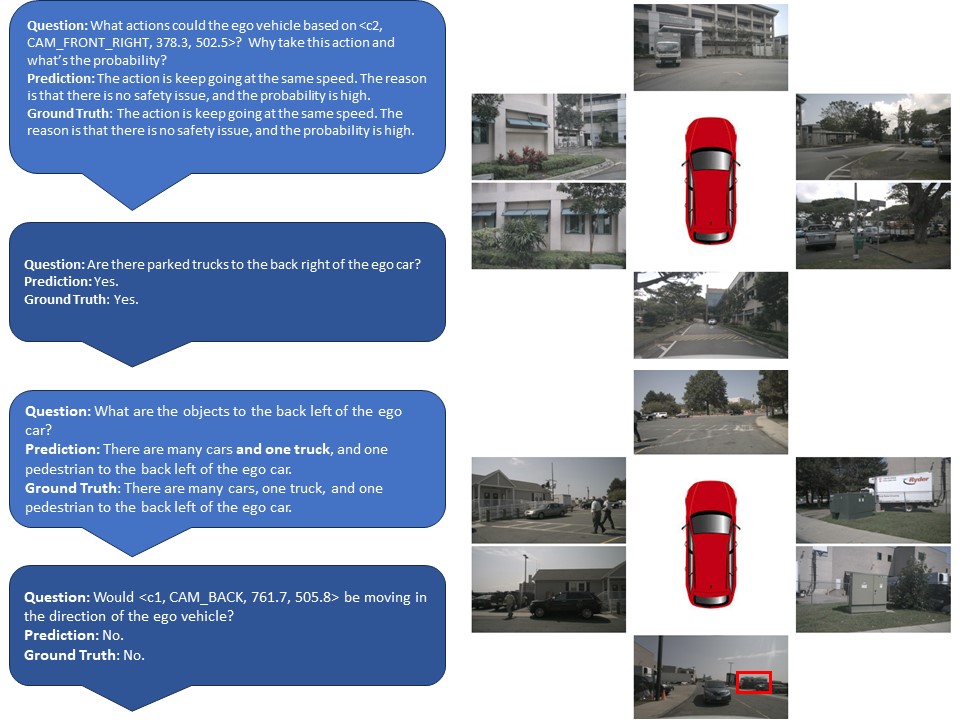}
    \vfill
    \includegraphics[width=0.8\textwidth]{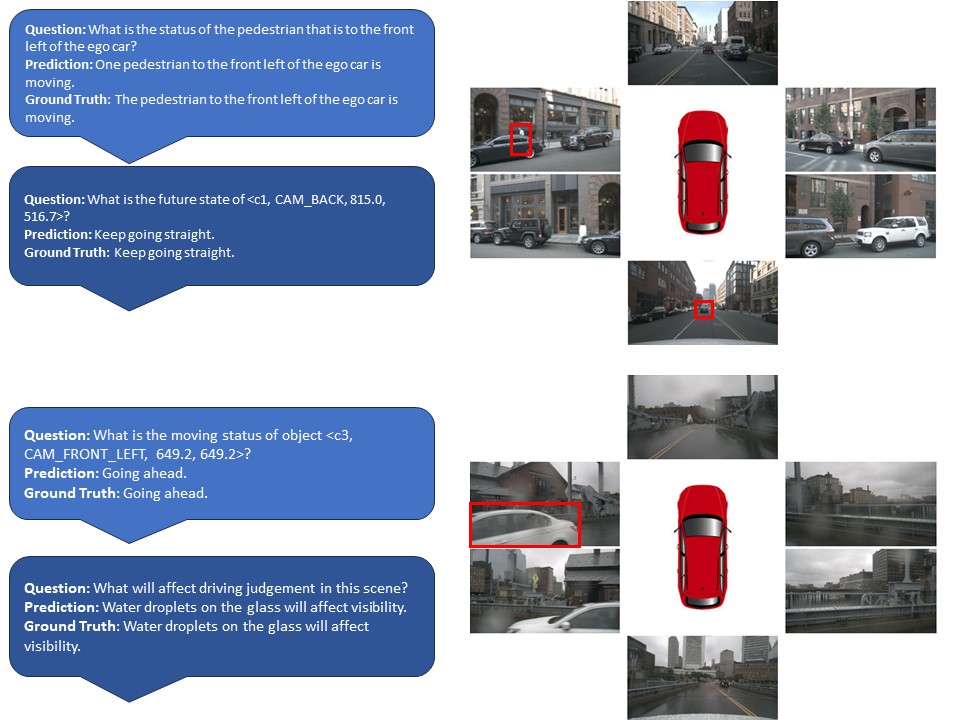}
    \vfill
    \caption{Example correct answer generations from EM-VLM4AD. As shown these in these examples, our model is able to perform VQA for various autonomous driving tasks such as perception, planning, and traffic agent behavior prediction.}
    \label{fig:DriveLM-generated}
\end{figure*}

\section{Methods}
Our model for Visual Question Answering (VQA) in Autonomous Driving, EM-VLM4AD, consists of a custom image embedding network and a pre-trained T5 language model \cite{raffel2023exploring}. We describe these following modules and the overall training process in this section.

\subsection{Image Embedding Network}
To tackle multi-view (Front, Front-Left, Front-Right, Back, Back-Left, Back-Right) QA tasks for autonomous driving, individual image embeddings need to be aggregated into a single embedding. This unified embedding can then be concatenated with a text embedding to serve as input to the LM. In typical VLMs, the image embedding process uses models like CLIP or object detection networks, resulting in a slow extraction process. To address this, we adopt the patch projection embedding scheme introduced by ViT \cite{dosovitskiy2020image}. Given an RGB image $I \in \mathbb{R}^{3 \times H \times W}$, the images are flattened and sliced into patches with a linear projection and positional embedding. This creates a latent image representation $V_i \in \mathbb{R}^{S_I \times H_I}$, where $S_I$ is the sequence length for the image embedding and $H_I$ is the hidden dimension of the image embedding. We use the pretrained weights of ViT-B/32 pretrained on ImageNet \cite{deng2009imagenet} to generate these image embeddings. 

This leaves us with 6 distinct individual image embeddings from each view, which now need to be combined. We first flatten each image embedding into a one-dimensional vector and then use gated pooling attention as described by Wu et al. \cite{wu2024mivc}. Given the individual image embeddings $V_i$, gated pooling attention learns a single embedding:
\begin{equation}
    V = \sum_{i=1}^N \alpha_i V_i
\end{equation}
in which $\alpha_i$ are weights for the ith image such that $\sum_{i=1}^N \alpha_i = 1$ that are calculated using:
\begin{equation}
    \alpha_i = \frac{exp\{w^T(tanh(ZV_i^T)\otimes sigm(GV_i^T)) \}}{\sum_{j=1}^N exp\{w^T(tanh(ZV_j^T)\otimes sigm(GV_j^T)) \}}
\end{equation}
where $w \in \mathbb{R}^K, Z \in \mathbb{R}^{K \times M}, G \in \mathbb{R}^{K \times M}, M = S_IH_I$, and $K$ is a hyperparameter we set to 128. Gated pooling attention introduces non-linearity which helps pool visual information across the image. With this combined image embedding $V \in \mathbb{R}^{S_I \times H_I}$, we then project this embedding to match the embedding dimension $H_T$ of the text embedding so that we can concatenate the text and image embedding together with dimension $\mathbb{R}^{(S_T + S_I) \times H_I}$, where $S_T$ is the sequence length of the text embedding. This concatenated, multimodal embedding is then inputted into the LM to generate answer text. 

\subsection{Language Model}
To mitigate the computational and inference costs of our vision-language model, we aim to use more lightweight LMs with less than one billion parameters. To achieve this, we use two different pre-trained versions of the T5 LM model: T5-Base, which contains around 223 million parameters, and an 8-bit quantized version of T5-Large ($\approx$ 750M parameters). Using these pre-trained LMs, we perform finetuning to adapt the LM to the concatenated multi-view image and text embeddings. In our experimentation, we found that fine-tuning the whole model for T5-Base works best, but for the quantized T5-Large we use LoRA-Fine-Tuning-Aware Quantization \cite{li2023loftq}, which helps minimize quantization error with the initialization of LoRA weights.  

\begin{figure*}[hbt!]
    \centering
    \includegraphics[width=0.8\textwidth]{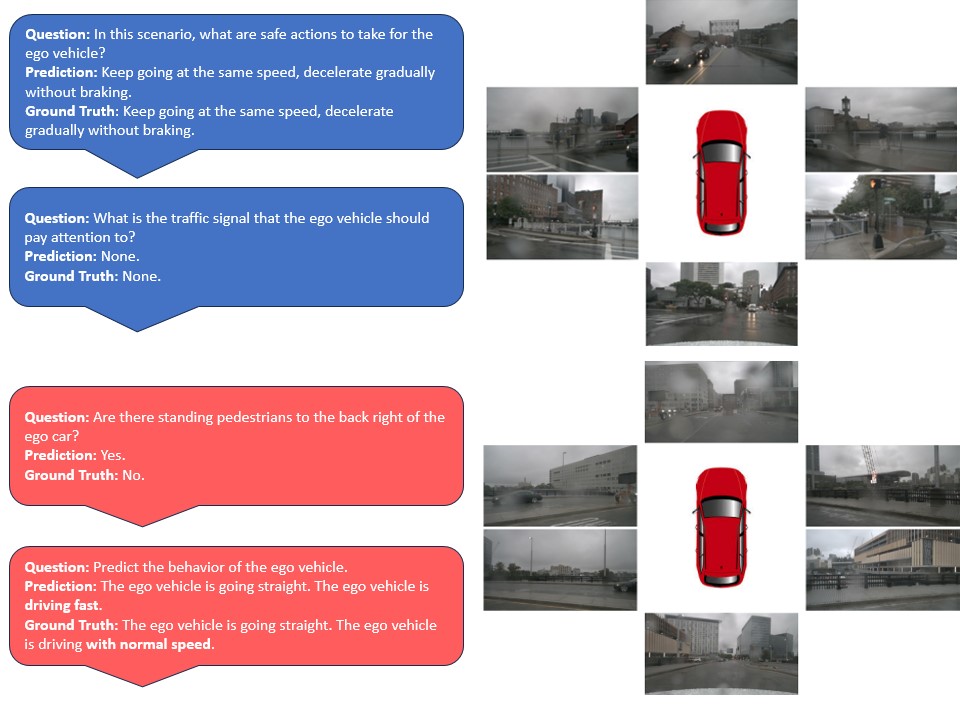}
    \caption{More example generations from EM-VLM4AD. As shown by the red QA examples, EM-VLM4AD can sometimes struggle with grammatical semantics and questions related to ego-vehicle behavior prediction, which may require video input for improved performance.}
    \label{fig:DriveLM-generated-2}
\end{figure*}

\subsection{Training Process}
To train EM-VLM4AD, we use the DriveLM dataset \cite{sima2023drivelm}, the most recent and comprehensive dataset for autonomous driving multi-view VQA with questions related to safety tasks such as perception, planning, prediction, and ego-vehicle behavior prediction. We use the training split of the DriveLM dataset, which contains 656 different scenes from NuScenes \cite{caesar2020nuscenes}, 4,072 different multi-view frames, and 377,983 different multi-view/QA pairs. To evaluate our approach, we use a 90\%/5\%/5\% split of the traffic scenes from DriveLM so we can evaluate how our model performs on unseen situations. Rather than train all components of our model in one stage, we use a two-stage approach as shown by Figure \ref{fig:DriveLM-examples}:

\begin{itemize}
    \item In the first stage, we first freeze the image patch encoder and the LM and only train the gated pooling attention and projection layer. This forces the multi-view image embeddings to align with the type of embeddings the LM expects.
    \item Then in the last stage, we only keep the image patch encoder frozen and start to finetune the LM. 
\end{itemize}

\begin{table*}[hbt!]
    \centering
    \begin{tabular}{|c|c|c|c|c|}
        \hline
         \textbf{Model} &  \textbf{BLEU-4 $\uparrow$} & \textbf{METEOR $\uparrow$} & \textbf{ROUGE-L $\uparrow$} & \textbf{CIDEr $\uparrow$}\\ 
         \hline
        $\text{EM-VLM4AD}_{\text{Base}}$ & 45.36 & 34.49 & \textbf{71.98} & \textbf{3.20} \\
        \hline
        $\text{EM-VLM4AD}_{\text{Q-Large}}$ & 40.11 & 34.34 & 70.72 & 3.10 \\
        \hline
        DriveLM-Agent \cite{sima2023drivelm} & \textbf{53.09} & \textbf{36.19} & 66.79 & 2.79 \\
        \hline
    \end{tabular}
    \caption{Qualitative comparison of generated answers between DriveLM-Agent and EM-VLM4AD on their respective test sets. $\text{EM-VLM4AD}_{\text{Base}}$ uses a T5-Base LM backbone, while $\text{EM-VLM4AD}_{\text{Q-Large}}$ uses an 8-bit quantized T5-Large backbone.}
    \label{tab:qual-results}
\end{table*}

In summary, the image patch encoder is always frozen to maintain generalized image information gathered from pretraining, the gated pooling attention and projection layer is always trained, and the Language Model is only finetuned during the last stage of training. \\
\indent We perform each training stage for six epochs, which takes around 2.5 days to finish for each model. We use a NVIDIA RTX 3090 Ti to train the T5-Large version of EM-VLM4AD and a V100 Google Colab instance to train EM-VLM4AD with T5-Base. We note that our models can be fit into a single T4 GPU instance, which allows to evaluate these models for free with Google Colab. For hyperparameters, we use a learning rate of 1e-4, weight decay of 0.05, an exponential learning rate scheduler, and a batch size of 4 for both approaches.

\section{Experiments}
This section presents an analysis of the quantitative, qualitative, and computational performance of EM-VLM4AD. We use the following metrics commonly used in image captioning tasks to assess the quality of the model-generated answers:
\begin{itemize}
    \item BLEU-4 \cite{papineni2002bleu}: Measures how many 4-grams in the generated text match those in the reference text.
    \item ROUGE-L \cite{lin2004rouge}: Calculates sentence similarity scores using the longest common sub-sequence between the generated text and ground-truth text.
    \item METEOR \cite{banerjee2005meteor}: Considers exact matches, stemming, synonymy, and word order to measure alignment between model outputs and references.
    \item CIDEr \cite{vedantam2015cider}: To account for lexical and semantic similarity between the generated and reference text, CIDEr weights n-grams with their corresponding TF-IDF weight. This helps de-emphasize n-grams that commonly occur across all examples that may not have important meaning.
\end{itemize}
For computational analysis, we aim to analyze the memory and computational efficiency of our model, essential aspects in real-time systems where resource constraints exist and inference efficiency is paramount.

\subsection{Quantitative Results}

We evaluate the BLEU-4, ROUGE-L, METEOR, and CIDEr scores using the test set of unseen traffic scenes we create. Currently, the only existing approach on the DriveLM dataset is DriveLM-Agent \cite{sima2023drivelm}, which is a finetuned version of BLIP-2. Since this model is not yet public and we do not have the computational resources to perform full-precision LoRA training of BLIP-2 , we benchmark our approach using the results Sima et al. \cite{sima2023drivelm} provide on their private evaluation set. The results from Table \ref{tab:qual-results} demonstrate how both versions of EM-VLM4AD outperform DriveLM-Agent on ROUGE-L and CIDEr, despite having at least 3 billion less model parameters. \\
\begin{table*}[hbt!]
    \centering
    \begin{tabular}{|c|C{3.7cm}|C{2.8cm}|c|c|c|}
        \hline
         \textbf{Model} &  \textbf{Pretrained Models Used} & \textbf{\# of Parameters $\downarrow$} & \textbf{FLOP Count $\downarrow$} & \textbf{Memory (GB) $\downarrow$}\\ 
         \hline
        $\text{EM-VLM4AD}_{\text{Base}}$ & T5-Base, ViT-b/32 patch embedder &  \textbf{235M} & \textbf{9.47B}  & 0.94 \\
        \hline
        $\text{EM-VLM4AD}_{\text{Q-Large}}$ & T5-Large, ViT-b/32 patch embedder & 769M & 31.5B & \textbf{0.77} \\
        \hline
        DriveLM-Agent \cite{sima2023drivelm} & BLIP-2 & 3.96B & 439B & 14.43 \\
        \hline
        DriveMLM  \cite{wang2023drivemlm} & LLaMA-7B, Vit-g/14 & 8.37B & 535B & 36 \\
        \hline
        LLM-Driver \cite{chen2023driving} & LLaMA-7B & 7B & 268B & 28 \\ 
        \hline
        Drive-GPT4 \cite{xu2023drivegpt4} & LLaMA 2, CLIP & 7.3B  & 329B  & 29.2 \\
        \hline
    \end{tabular}
    \caption{Computational comparison of other LMs for Autonomous Driving with both versions of EM-VLM4AD. The EM-VLM4AD models have the smallest number of parameters, memory space, and FLOP count, making them the most efficient and computationally efficient VLM for autonomous driving.}
    \label{tab:comp-results}
\end{table*}
\indent The superior performance of EM-VLM4AD with the T5-Base backbone over the 8-bit quantized T5-Large version can be attributed to the former's ability to train a larger parameter set. This facilitates a better adaptation of the language model to the input vision-language embeddings. Conversely, the LoRA finetuning approach for the 8-bit quantized T5-Large LM only changes 3.4\% of the network's weights. While we did try full finetuning for the quantized LM, this over fine-tuned the LM and caused mode collapse. \\
\indent The integration of multiple frames is a critical advantage that contributes to EM-VLM4AD's performance versus DriveLM-Agent. Unlike DriveLM-Agent, which only uses the front-view frame as input, our model successfully aggregates information across multiple views with our custom multi-view embedding network. Furthermore, while certain tasks done by LMs are defined as \emph{emergent}, requiring larger models for sufficient results, our study underscores that learning to perform VQA on the DriveLM dataset can be done without increasing model complexity. Therefore, simply adding model complexity may not result in optimal improvements for this specific task.

\subsection{Computational Analysis}

We also perform computational analysis to see how EM-VLM4AD compares to other multimodal LMs for autonomous driving. Specifically, we focus on three key computational metrics: the \# of parameters, \# of Floating Point Operations (FLOPs), and memory in gigabytes (GB). For these methods, the image encoder and LM constitute the most computationally expensive aspects of these models, so we only focus on these two aspects when calculating these metrics. To estimate the FLOP count for each of these models, we use the \href{https://github.com/facebookresearch/fvcore/blob/main/docs/flop_count.md}{fvcore FLOP counter} module on examples from the DriveLM dataset with a A100 GPU. For the methods we compare to, we add the FLOPs of the image encoder and LM together. The results in Table \ref{tab:comp-results} underscore that EM-VLM4AD is considerably more efficient than other methods, requiring less memory, computations, and model parameters. Notably, EM-VLM4AD with the T5-Base backbone has the least parameters and FLOP count, while EM-VLM4AD with the T5-Large backbone has the least memory requirements since model weights are only stored in 8 bits. These optimized model design choices enable EM-VLM4AD to provide fast inference times and require less computational resources, critical attributes for any LM implemented for real-time scenarios.

\subsection{Qualitative Results}
Figures \ref{fig:DriveLM-generated} and \ref{fig:DriveLM-generated-2} showcase some selected multi-frame answer generations produced by EM-VLM4AD. Our model can accurately respond to a variety of questions related to perception, traffic agent behavior identification, planning safe ego-vehicle behavior, and identifying important traffic elements in a scene. Through leveraging the general knowledge from the pretrained patch embedding network and T5-LM, our system can answer a wide spectrum of questions that encapsulate an end-to-end autonomous driving system. Additionally, EM-VLM4AD demonstrates the ability to understand the c-tag format employed by DriveLM, which encodes traffic objects as $< c, CAM, x_{\text{pos}}, y_{\text{pos}}>$. Moreover, this model learns to intelligently extract the most relevant frames for each question, making it an effective multi-frame VLM system. However, EM-VLM4AD exhibits one specific weaknesses: answering questions related to behavior where the prompt is ``Predict the behavior for the ego vehicle". 
% EM-VLM4AD can occasionally generate answers with grammatical errors, hindering someone to understand the answer to a question. Adding training techniques such as distillation \cite{hinton2015distilling} with larger vision-language models, which have a better understanding of grammar rules, will help this smaller model learn these complex rules. 
Adding temporal context through inputting multi-view videos to our network would improve results on this type of question, since behavior related questions often need more than one frame to make accurate predictions.

\section{Conclusion}
We introduce EM-VLM4AD, a lightweight multi-frame vision-language model designed for Visual Question Answering across various autonomous driving tasks. Compared to other LMs tailored for autonomous driving, EM-VLM4AD exhibits notable advantages in terms of memory efficiency and computational requirements, and outperforms the reported scores of DriveLM-Agent in ROUGE and CIDEr metrics on a DriveLM test dataset. EM-VLM4AD demonstrates proficiency in responding to a variety of autonomous driving questions and dynamically focuses on relevant camera views through our gated pooling attention layer, which effectively integrates view embeddings. 
In future research, we aspire to evolve our model into a video-language model capable of generating responses from multi-view video inputs, thereby enhancing EM-VLM4AD's ability to handle temporal-related inquiries. Furthermore, incorporating multimodal retrieval augmented generation to provide context can enable our model to extract insights from analogous traffic scenarios.
{
    \small
    \bibliographystyle{ieeenat_fullname}
    \bibliography{main}

\begin{thebibliography}{37}
\providecommand{\natexlab}[1]{#1}
\providecommand{\url}[1]{\texttt{#1}}
\expandafter\ifx\csname urlstyle\endcsname\relax
  \providecommand{\doi}[1]{doi: #1}\else
  \providecommand{\doi}{doi: \begingroup \urlstyle{rm}\Url}\fi

\bibitem[Alayrac et~al.(2022)Alayrac, Donahue, Luc, Miech, Barr, Hasson, Lenc, Mensch, Millican, Reynolds, et~al.]{alayrac2022flamingo}
Jean-Baptiste Alayrac, Jeff Donahue, Pauline Luc, Antoine Miech, Iain Barr, Yana Hasson, Karel Lenc, Arthur Mensch, Katherine Millican, Malcolm Reynolds, et~al.
\newblock Flamingo: a visual language model for few-shot learning.
\newblock \emph{Advances in Neural Information Processing Systems}, 35:\penalty0 23716--23736, 2022.

\bibitem[Almazrouei et~al.(2023)Almazrouei, Alobeidli, Alshamsi, Cappelli, Cojocaru, Debbah, Étienne Goffinet, Hesslow, Launay, Malartic, Mazzotta, Noune, Pannier, and Penedo]{almazrouei2023falcon}
Ebtesam Almazrouei, Hamza Alobeidli, Abdulaziz Alshamsi, Alessandro Cappelli, Ruxandra Cojocaru, Mérouane Debbah, Étienne Goffinet, Daniel Hesslow, Julien Launay, Quentin Malartic, Daniele Mazzotta, Badreddine Noune, Baptiste Pannier, and Guilherme Penedo.
\newblock The falcon series of open language models, 2023.

\bibitem[Banerjee and Lavie(2005)]{banerjee2005meteor}
Satanjeev Banerjee and Alon Lavie.
\newblock Meteor: An automatic metric for mt evaluation with improved correlation with human judgments.
\newblock In \emph{Proceedings of the acl workshop on intrinsic and extrinsic evaluation measures for machine translation and/or summarization}, pages 65--72, 2005.

\bibitem[Caesar et~al.(2020)Caesar, Bankiti, Lang, Vora, Liong, Xu, Krishnan, Pan, Baldan, and Beijbom]{caesar2020nuscenes}
Holger Caesar, Varun Bankiti, Alex~H Lang, Sourabh Vora, Venice~Erin Liong, Qiang Xu, Anush Krishnan, Yu Pan, Giancarlo Baldan, and Oscar Beijbom.
\newblock nuscenes: A multimodal dataset for autonomous driving.
\newblock In \emph{Proceedings of the IEEE/CVF conference on computer vision and pattern recognition}, pages 11621--11631, 2020.

\bibitem[Chen et~al.(2023)Chen, Sinavski, H{\"u}nermann, Karnsund, Willmott, Birch, Maund, and Shotton]{chen2023driving}
Long Chen, Oleg Sinavski, Jan H{\"u}nermann, Alice Karnsund, Andrew~James Willmott, Danny Birch, Daniel Maund, and Jamie Shotton.
\newblock Driving with llms: Fusing object-level vector modality for explainable autonomous driving.
\newblock \emph{arXiv preprint arXiv:2310.01957}, 2023.

\bibitem[Cho et~al.(2021)Cho, Lei, Tan, and Bansal]{cho2021unifying}
Jaemin Cho, Jie Lei, Hao Tan, and Mohit Bansal.
\newblock Unifying vision-and-language tasks via text generation.
\newblock In \emph{International Conference on Machine Learning}, pages 1931--1942. PMLR, 2021.

\bibitem[Dai et~al.(2023)Dai, Li, Li, Tiong, Zhao, Wang, Li, Fung, and Hoi]{dai2023instructblip}
Wenliang Dai, Junnan Li, Dongxu Li, Anthony Meng~Huat Tiong, Junqi Zhao, Weisheng Wang, Boyang Li, Pascale Fung, and Steven Hoi.
\newblock Instructblip: Towards general-purpose vision-language models with instruction tuning, 2023.

\bibitem[Deng et~al.(2009)Deng, Dong, Socher, Li, Li, and Fei-Fei]{deng2009imagenet}
Jia Deng, Wei Dong, Richard Socher, Li-Jia Li, Kai Li, and Li Fei-Fei.
\newblock Imagenet: A large-scale hierarchical image database.
\newblock In \emph{2009 IEEE conference on computer vision and pattern recognition}, pages 248--255. Ieee, 2009.

\bibitem[Devlin et~al.(2018)Devlin, Chang, Lee, and Toutanova]{devlin2018bert}
Jacob Devlin, Ming-Wei Chang, Kenton Lee, and Kristina Toutanova.
\newblock Bert: Pre-training of deep bidirectional transformers for language understanding.
\newblock \emph{arXiv preprint arXiv:1810.04805}, 2018.

\bibitem[Dosovitskiy et~al.(2020)Dosovitskiy, Beyer, Kolesnikov, Weissenborn, Zhai, Unterthiner, Dehghani, Minderer, Heigold, Gelly, et~al.]{dosovitskiy2020image}
Alexey Dosovitskiy, Lucas Beyer, Alexander Kolesnikov, Dirk Weissenborn, Xiaohua Zhai, Thomas Unterthiner, Mostafa Dehghani, Matthias Minderer, Georg Heigold, Sylvain Gelly, et~al.
\newblock An image is worth 16x16 words: Transformers for image recognition at scale.
\newblock \emph{arXiv preprint arXiv:2010.11929}, 2020.

\bibitem[Du et~al.(2022)Du, Liu, Li, and Zhao]{du2022survey}
Yifan Du, Zikang Liu, Junyi Li, and Wayne~Xin Zhao.
\newblock A survey of vision-language pre-trained models.
\newblock \emph{arXiv preprint arXiv:2202.10936}, 2022.

\bibitem[Fang et~al.(2021)Fang, Wang, Hu, Wang, Yang, and Liu]{fang2021compressing}
Zhiyuan Fang, Jianfeng Wang, Xiaowei Hu, Lijuan Wang, Yezhou Yang, and Zicheng Liu.
\newblock Compressing visual-linguistic model via knowledge distillation.
\newblock In \emph{Proceedings of the IEEE/CVF International Conference on Computer Vision}, pages 1428--1438, 2021.

\bibitem[Greer and Trivedi(2024)]{greer2024towards}
Ross Greer and Mohan Trivedi.
\newblock Towards explainable, safe autonomous driving with language embeddings for novelty identification and active learning: Framework and experimental analysis with real-world data sets.
\newblock \emph{arXiv preprint arXiv:2402.07320}, 2024.

\bibitem[Greer et~al.(2023)Greer, Gopalkrishnan, Landgren, Rakla, Gopalan, and Trivedi]{greer2023robust}
Ross Greer, Akshay Gopalkrishnan, Jacob Landgren, Lulua Rakla, Anish Gopalan, and Mohan Trivedi.
\newblock Robust traffic light detection using salience-sensitive loss: Computational framework and evaluations.
\newblock In \emph{2023 IEEE Intelligent Vehicles Symposium (IV)}, pages 1--7. IEEE, 2023.

\bibitem[Greer et~al.(2024{\natexlab{a}})Greer, Antoniussen, Andersen, M{\o}gelmose, and Trivedi]{greer2024and}
Ross Greer, Bj{\o}rk Antoniussen, Mathias~V Andersen, Andreas M{\o}gelmose, and Mohan~M Trivedi.
\newblock The why, when, and how to use active learning in large-data-driven 3d object detection for safe autonomous driving: An empirical exploration.
\newblock \emph{arXiv preprint arXiv:2401.16634}, 2024{\natexlab{a}}.

\bibitem[Greer et~al.(2024{\natexlab{b}})Greer, Gopalkrishnan, Keskar, and Trivedi]{greer2024patterns}
Ross Greer, Akshay Gopalkrishnan, Maitrayee Keskar, and Mohan~M Trivedi.
\newblock Patterns of vehicle lights: Addressing complexities of camera-based vehicle light datasets and metrics.
\newblock \emph{Pattern Recognition Letters}, 178:\penalty0 209--215, 2024{\natexlab{b}}.

\bibitem[Hu et~al.(2021)Hu, Shen, Wallis, Allen-Zhu, Li, Wang, Wang, and Chen]{hu2021lora}
Edward~J Hu, Yelong Shen, Phillip Wallis, Zeyuan Allen-Zhu, Yuanzhi Li, Shean Wang, Lu Wang, and Weizhu Chen.
\newblock Lora: Low-rank adaptation of large language models.
\newblock \emph{arXiv preprint arXiv:2106.09685}, 2021.

\bibitem[Kim et~al.(2018)Kim, Rohrbach, Darrell, Canny, and Akata]{kim2018textual}
Jinkyu Kim, Anna Rohrbach, Trevor Darrell, John Canny, and Zeynep Akata.
\newblock Textual explanations for self-driving vehicles.
\newblock In \emph{Proceedings of the European conference on computer vision (ECCV)}, pages 563--578, 2018.

\bibitem[Li et~al.(2023{\natexlab{a}})Li, Li, Savarese, and Hoi]{li2023blip}
Junnan Li, Dongxu Li, Silvio Savarese, and Steven Hoi.
\newblock Blip-2: Bootstrapping language-image pre-training with frozen image encoders and large language models.
\newblock \emph{arXiv preprint arXiv:2301.12597}, 2023{\natexlab{a}}.

\bibitem[Li et~al.(2023{\natexlab{b}})Li, Fang, Liu, Ling, Tu, and Su]{li2023distilling}
Xuanlin Li, Yunhao Fang, Minghua Liu, Zhan Ling, Zhuowen Tu, and Hao Su.
\newblock Distilling large vision-language model with out-of-distribution generalizability.
\newblock In \emph{Proceedings of the IEEE/CVF International Conference on Computer Vision}, pages 2492--2503, 2023{\natexlab{b}}.

\bibitem[Li et~al.(2023{\natexlab{c}})Li, Yu, Liang, He, Karampatziakis, Chen, and Zhao]{li2023loftq}
Yixiao Li, Yifan Yu, Chen Liang, Pengcheng He, Nikos Karampatziakis, Weizhu Chen, and Tuo Zhao.
\newblock Loftq: Lora-fine-tuning-aware quantization for large language models, 2023{\natexlab{c}}.

\bibitem[Lin(2004)]{lin2004rouge}
Chin-Yew Lin.
\newblock Rouge: A package for automatic evaluation of summaries.
\newblock In \emph{Text summarization branches out}, pages 74--81, 2004.

\bibitem[Mao et~al.(2023)Mao, Qian, Zhao, and Wang]{mao2023gpt}
Jiageng Mao, Yuxi Qian, Hang Zhao, and Yue Wang.
\newblock Gpt-driver: Learning to drive with gpt.
\newblock \emph{arXiv preprint arXiv:2310.01415}, 2023.

\bibitem[Messaoud et~al.(2021)Messaoud, Deo, Trivedi, and Nashashibi]{messaoud2021trajectory}
Kaouther Messaoud, Nachiket Deo, Mohan~M Trivedi, and Fawzi Nashashibi.
\newblock Trajectory prediction for autonomous driving based on multi-head attention with joint agent-map representation.
\newblock In \emph{2021 IEEE Intelligent Vehicles Symposium (IV)}, pages 165--170. IEEE, 2021.

\bibitem[Papineni et~al.(2002)Papineni, Roukos, Ward, and Zhu]{papineni2002bleu}
Kishore Papineni, Salim Roukos, Todd Ward, and Wei-Jing Zhu.
\newblock Bleu: a method for automatic evaluation of machine translation.
\newblock In \emph{Proceedings of the 40th annual meeting of the Association for Computational Linguistics}, pages 311--318, 2002.

\bibitem[Radford et~al.(2019)Radford, Wu, Child, Luan, Amodei, Sutskever, et~al.]{radford2019language}
Alec Radford, Jeffrey Wu, Rewon Child, David Luan, Dario Amodei, Ilya Sutskever, et~al.
\newblock Language models are unsupervised multitask learners.
\newblock \emph{OpenAI blog}, 1\penalty0 (8):\penalty0 9, 2019.

\bibitem[Radford et~al.(2021)Radford, Kim, Hallacy, Ramesh, Goh, Agarwal, Sastry, Askell, Mishkin, Clark, et~al.]{radford2021learning}
Alec Radford, Jong~Wook Kim, Chris Hallacy, Aditya Ramesh, Gabriel Goh, Sandhini Agarwal, Girish Sastry, Amanda Askell, Pamela Mishkin, Jack Clark, et~al.
\newblock Learning transferable visual models from natural language supervision.
\newblock In \emph{International conference on machine learning}, pages 8748--8763. PMLR, 2021.

\bibitem[Raffel et~al.(2023)Raffel, Shazeer, Roberts, Lee, Narang, Matena, Zhou, Li, and Liu]{raffel2023exploring}
Colin Raffel, Noam Shazeer, Adam Roberts, Katherine Lee, Sharan Narang, Michael Matena, Yanqi Zhou, Wei Li, and Peter~J. Liu.
\newblock Exploring the limits of transfer learning with a unified text-to-text transformer, 2023.

\bibitem[Sha et~al.(2023)Sha, Mu, Jiang, Chen, Xu, Luo, Li, Tomizuka, Zhan, and Ding]{sha2023languagempc}
Hao Sha, Yao Mu, Yuxuan Jiang, Li Chen, Chenfeng Xu, Ping Luo, Shengbo~Eben Li, Masayoshi Tomizuka, Wei Zhan, and Mingyu Ding.
\newblock Languagempc: Large language models as decision makers for autonomous driving, 2023.

\bibitem[Sima et~al.(2023)Sima, Renz, Chitta, Chen, Zhang, Xie, Luo, Geiger, and Li]{sima2023drivelm}
Chonghao Sima, Katrin Renz, Kashyap Chitta, Li Chen, Hanxue Zhang, Chengen Xie, Ping Luo, Andreas Geiger, and Hongyang Li.
\newblock Drivelm: Driving with graph visual question answering.
\newblock \emph{arXiv preprint arXiv:2312.14150}, 2023.

\bibitem[Touvron et~al.(2023)Touvron, Lavril, Izacard, Martinet, Lachaux, Lacroix, Rozi{\`e}re, Goyal, Hambro, Azhar, et~al.]{touvron2023llama}
Hugo Touvron, Thibaut Lavril, Gautier Izacard, Xavier Martinet, Marie-Anne Lachaux, Timoth{\'e}e Lacroix, Baptiste Rozi{\`e}re, Naman Goyal, Eric Hambro, Faisal Azhar, et~al.
\newblock Llama: Open and efficient foundation language models.
\newblock \emph{arXiv preprint arXiv:2302.13971}, 2023.

\bibitem[Vaswani et~al.(2017)Vaswani, Shazeer, Parmar, Uszkoreit, Jones, Gomez, Kaiser, and Polosukhin]{vaswani2017attention}
Ashish Vaswani, Noam Shazeer, Niki Parmar, Jakob Uszkoreit, Llion Jones, Aidan~N Gomez, {\L}ukasz Kaiser, and Illia Polosukhin.
\newblock Attention is all you need.
\newblock \emph{Advances in neural information processing systems}, 30, 2017.

\bibitem[Vedantam et~al.(2015)Vedantam, Lawrence~Zitnick, and Parikh]{vedantam2015cider}
Ramakrishna Vedantam, C Lawrence~Zitnick, and Devi Parikh.
\newblock Cider: Consensus-based image description evaluation.
\newblock In \emph{Proceedings of the IEEE conference on computer vision and pattern recognition}, pages 4566--4575, 2015.

\bibitem[Wang et~al.(2023)Wang, Xie, Hu, Zou, Fan, Tong, Wen, Wu, Deng, Li, Tian, Lu, Zhu, Wang, Qiao, and Dai]{wang2023drivemlm}
Wenhai Wang, Jiangwei Xie, ChuanYang Hu, Haoming Zou, Jianan Fan, Wenwen Tong, Yang Wen, Silei Wu, Hanming Deng, Zhiqi Li, Hao Tian, Lewei Lu, Xizhou Zhu, Xiaogang Wang, Yu Qiao, and Jifeng Dai.
\newblock Drivemlm: Aligning multi-modal large language models with behavioral planning states for autonomous driving, 2023.

\bibitem[Wei et~al.(2023)Wei, Wang, Schuurmans, Bosma, Ichter, Xia, Chi, Le, and Zhou]{wei2023chainofthought}
Jason Wei, Xuezhi Wang, Dale Schuurmans, Maarten Bosma, Brian Ichter, Fei Xia, Ed Chi, Quoc Le, and Denny Zhou.
\newblock Chain-of-thought prompting elicits reasoning in large language models, 2023.

\bibitem[Wu et~al.(2024)Wu, Li, Zhong, and Huang]{wu2024mivc}
Wenyi Wu, Qi Li, Wenliang Zhong, and Junzhou Huang.
\newblock Mivc: Multiple instance visual component for visual-language models.
\newblock In \emph{Proceedings of the IEEE/CVF Winter Conference on Applications of Computer Vision}, pages 8117--8126, 2024.

\bibitem[Xu et~al.(2023)Xu, Zhang, Xie, Zhao, Guo, Wong, Li, and Zhao]{xu2023drivegpt4}
Zhenhua Xu, Yujia Zhang, Enze Xie, Zhen Zhao, Yong Guo, Kenneth~KY Wong, Zhenguo Li, and Hengshuang Zhao.
\newblock Drivegpt4: Interpretable end-to-end autonomous driving via large language model.
\newblock \emph{arXiv preprint arXiv:2310.01412}, 2023.

\end{thebibliography}
}


\end{document}